\documentclass[a4paper,11pt]{article}

\usepackage{graphicx}  %%% for including graphics
\usepackage{url}       %%% for including URLs
\usepackage{times}
\usepackage{natbib}
\usepackage[margin=25mm]{geometry}
\usepackage{url}
\usepackage{booktabs}
\usepackage{microtype}
\usepackage{amsmath,amsfonts,amssymb,mathtools,bm}
\usepackage{graphicx}
\usepackage{paralist}

\DeclareMathOperator*{\softmax}{softmax}
\title{Living a discrete life in a continuous world: \\ 
  Reference in cross-modal entity tracking}
\date{}

\author{
Gemma Boleda$^1$ \ Sebastian Pad\'o$^2$ \ Nghia The Pham$^3$ \ Marco
Baroni$^4$\\
$^1$Universitat Pompeu Fabra\\[2pt]
$^2$Institut f\"ur Maschinelle Sprachverarbeitung,
Universit\"at Stuttgart\\[2pt]
$^3$Center for Mind/Brain Sciences,
University of Trento\\[2pt]
$^4$Facebook Artificial Intelligence Research,
Paris\\
{\tt gemma.boleda@upf.edu}, {\tt pado@ims.uni-stuttgart.de},\\
{\tt thenghia.pham@unitn.it}, {\tt mbaroni@fb.com}
\\}

\begin{document}
\maketitle
\thispagestyle{empty}
\pagestyle{empty}

\begin{abstract}
  \noindent Reference is a crucial property of language that allows us
  to connect linguistic expressions to the world. Modeling it requires
  handling both continuous and discrete aspects of
  meaning. Data-driven models excel at the former, but struggle with
  the latter, and the reverse is true for symbolic models.

  This paper (a)~introduces a concrete referential task to test both
  aspects, called cross-modal entity tracking; (b)~proposes a neural
  network architecture that uses external memory to build an
  \textit{entity library} inspired in the DRSs of DRT, with a
  mechanism to dynamically introduce new referents or add information
  to referents that are already in the library.

  Our model shows promise: it beats traditional neural network
  architectures on the task. However, it is still outperformed by
  Memory Networks, another model with external memory.
\end{abstract}

\section{Introduction}
\label{sec:intro}

Language combines discrete and continuous facets, as exemplified by
the phenomenon of \textit{reference}~\citep{frege1892,abbott10}: When
we refer to an object in the world with the noun phrase \textit{the
  mug I bought}, we use content words such as \textit{mug}, which are
notoriously fuzzy or vague in their
meaning~\citep{van-deemter12,murphy02} and are best modeled through
continuous means~\citep{boleda-herbelot16}. Once the referent for the
mug has been established, however, it becomes a linguistic entity that
we can manipulate in a largely discrete fashion, retrieving it and
updating it with new information as needed \citep[\textit{Remember the
    mug I bought? My brother stole it!}][]{kamp-reyle93}. Put
  differently, managing reference requires two distinct abilities:
\begin{compactenum}
\item The ability to \emph{categorize}, that is, to recognize that
  different entities are equivalent with regard to some concept of
  interest \citep[e.g.\ two mugs, two instances of the ``things to
    take on a camping trip'' category;][]{Barsalou1983}. This implies
  being able to aggregate seemingly diverse objects.
\item The ability to \emph{individuate}, that is, to keep entities
  distinct even if they are similar with regard to many attributes
  (e.g.\ two pieces of pink granite that were collected in different
  national parks). This implies being able to keep seemingly similar
  things apart.
\end{compactenum}

Data-driven, continuous models are very good at categorizing, but not
at individuating, and the reverse holds for symbolic
models~\citep{boleda-herbelot16}. Our long-term research goal is to
build a \textbf{continuous computational model of reference} that
emulates discrete referential mechanisms such as those defined in DRT
\citep{kamp-reyle93}; here we present initial work towards that goal,
with two specific contributions.

Our first contribution is an experimental task (and associated
dataset), \textbf{cross-modal entity tracking}, that tests the ability
of computational models to refer successfully in a setting where they
are required to both categorize and individuate entities. The task
presents different entities (represented by pictures) repeatedly, each
time with a different, linguistically conveyed attribute (e.g.\ a
given mug is presented once with the attribute \textit{bought} and
once with \textit{stolen}). The category label (``mug'') is not given
at exposure time. The task is to choose the picture of the entity that
corresponds to a linguistic query that combines category information
with attribute information (e.g.\ simulating ``the mug that was bought
and stolen''), among the set of all the entities presented in a given
sequence. The sequences in each datapoint of our dataset contain
confounders that make the task challenging: Other entities with the
same category but only one matching attribute (e.g.\ a different mug
that was bought and stored), and other entities with the same
attributes but a different category (e.g.\ a chair that was bought and
stolen). Therefore, the task requires models to 1)~correctly
categorize entities, recognizing which images belong to the category
in the query (something that is hard for symbolic models),~2)
individuate and track them, being able to distinguish among different
entities based on visual and linguistic cues provided at different
time steps (something that is hard for continuous models).

In DRT terms \citep{kamp-reyle93}, each entity exposure either
introduces a new discourse referent or updates the representation of
an old referent with new information. To solve the task successfully,
the model needs to decide, for each incoming exposure, whether to
aggregate it with a previously known referent (in DRT, this means
introducing an equation between two referents), or to treat it as a
new referent.

Our second contribution is a neural network architecture with a module
for referent representations: \textbf{DIstributed model of REference},
\textbf{DIRE}. DIRE uses the concept of \textit{external memory} from
deep learning \citep{joulin-mikolov15,graves+16} to build an entity
library for an exposure sequence that conceptually corresponds to the
set of DRT discourse referents, using similarity-based reasoning on
distributed representations to decide between aggregating and
initializing entity representations. In contrast to symbolic
implementations of DRT~\citep{bos08}, which manipulate discourse
referents on the basis of manually specified algorithms, DIRE learns
to make these decisions directly from observing reference acts using
end-to-end training. We see our paper as a first, modest step in the
direction of data-driven learning of DRT-like behavior, and are of
course still far from learning anything resembling a fully fledged DRT
system.

\section{Cross-modal Entity Tracking: Task and Data}
\label{sec:data}

\paragraph{Task.}
Imagine an office, with a desk where there are three mugs and other
objects. Adam tells Barbara that he just bought two of the mugs and he
particularly likes the one on the right. Later they are in the
kitchen, and Adam, busy preparing coffee, asks Barbara: ``Remember the
mugs I bought? Could you please bring the one I like?''. To pick the
right mug from the office, Barbara must correctly categorize the
objects on the desk (identify which ones are mugs) and individuate
them via their properties (singling out the one Adam is asking for).
Also, she must combine visual and linguistically conveyed properties
of the objects: Visual properties tell her which ones are mugs, the
properties that Adam told her about help her pick the right one. Our
\textit{cross-modal entity tracking task} emulates this kind of
situation. Our current study uses a simplified version of the task
that allows us to carefully control all the variables involved.

\begin{figure}[t]
  \centering
  \includegraphics[width=0.8\columnwidth]{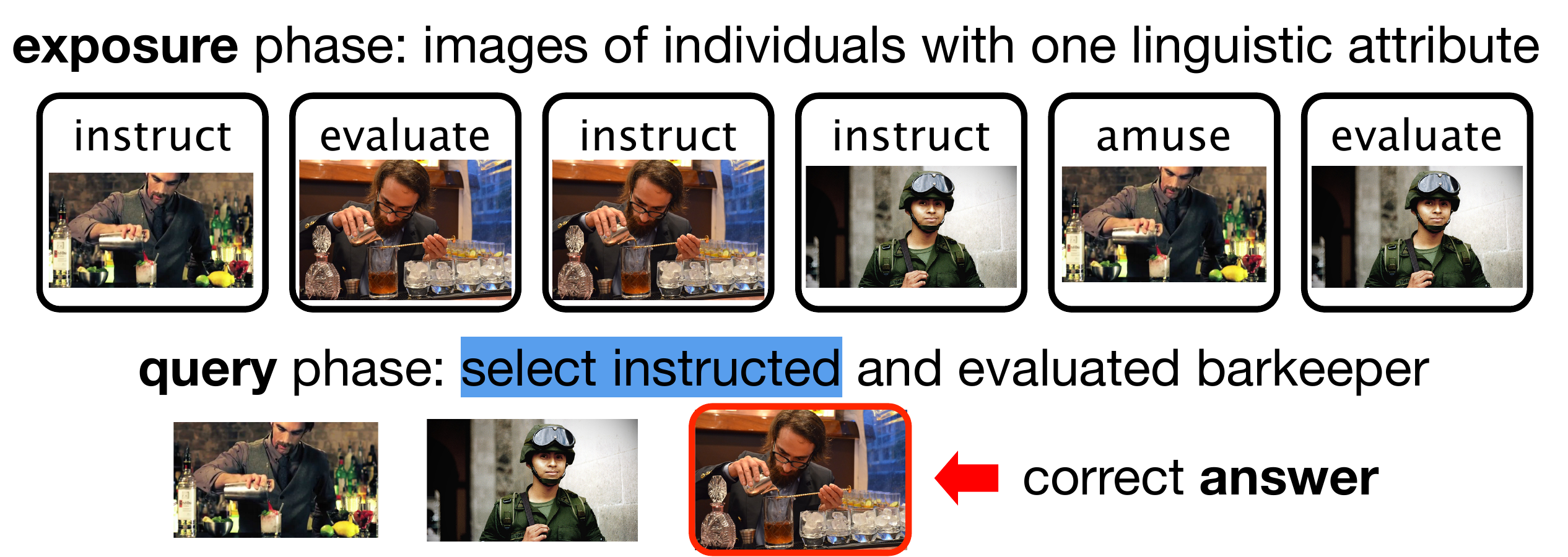}
  \vspace*{-1.5em}
  \caption{Cross-modal tracking task (actual datapoints
    contain 12 exposures and 6 images to pick from).}
  \label{fig:tracking-example}
\end{figure}

We operationalize the task as one of pointing to real-life pictures of
objects. Figure~\ref{fig:tracking-example} shows a simplified
example. We sample six entities belonging to two categories (in the
example, where only three entities are shown, barkeepers and
soldiers). Each entity is represented by one image (that is, barkeeper
A is always represented by the same image). We also sample different
attributes, which are compatible with both categories (in the example,
``instructed'', ``evaluated'', ``amused''). In the exposure phase, we
present each entity (image) twice at different time steps, each time
with one of the attributes. In this phase, the category of the entity
in the image is not given to the model, only the images are. At query
time, we use a linguistic query with one category (e.g.,
``barkeeper'') and two attributes (e.g., ``instructed and
evaluated''). The task is to retrieve the image of the entity that
corresponds to the query. To solve it, it is not enough to rely on
categorization or object labeling (in the actual task, there are
always three entities belonging to the category in the query), nor is
it enough to rely on attribute information (there will always be three
entities for each attribute, and two for the combination of attributes
in the query). Note that one important simplification we make, with
respect to a real-life scenario, is that an entity is always
represented by the very same image. The current setup is nevertheless
already very challenging for current models, as the experiments below
will show. Indeed, to succeed in the task a model must correctly
associate the category in the query with images of the right object,
it must develop a mechanism to index entities based on the images
representing them, and it must learn to correctly accumulate over time
the different attributes to be stored with each entity.

% In the exposure
% phase, the model processes a sequence of \textit{exposures}, each
% consisting of the image of an object together with one linguistic
% attribute that cannot be inferred from the image (e.g., the attribute
% \textit{instructed} for the first barkeeper). % \footnote{We
%  %  use the base form of verbs (e.g.\ \textit{instruct}) for
%  %  convenience. The example shows just three entities of two
%  %  categories. See below for information on our actual dataset.}
%  % % 
% Note that the object category (\textit{barkeeper}, \textit{soldier})
% is not given. We present the same entities with different attributes
% at different time steps (e.g., the first barkeeper reappears with the
% attribute \textit{amused}).  In the query phase, the model is
% presented with a purely linguistic query involving two attributes and
% an object (e.g., \textit{instructed and evaluated barkeeper}), plus
% the set of all unique images presented before. The model must select
% the one image that satisfies the query. As (partially) shown in
% Figure~\ref{fig:tracking-example}, the input sequence contains
% distractors: A soldier sharing both attributes, a different barkeeper
% sharing only one of the attributes. The model must construct and
% update entity representations and keep them distinct based on their
% attributes (both visual and non-visual), addressing the three
% challenges listed above. Also, the model must handle both continuous
% aspects (categorization, by mapping across modalities to match images
% with nouns) and discrete aspects (identifying the one entity that
% matches the query).

The task is related to coreference resolution~\citep[see][for a recent
  survey]{poesio+17:anaphora_book}, but focuses on identifying
language-external objects from images rather than mentions of a
referent in text; to Visual Question Answering~\citep{antol+15}, but
it cannot be solved with visual information alone; and to Referring
Expression Generation~\citep{krahmer-van-deemter12}, but involves
identification rather than generation.

\paragraph{Dataset.} We have constructed a dataset for the task
containing 40k sequences for training, 5k for validation and 10k for
testing.\footnote{Available at
  \url{http://www.ims.uni-stuttgart.de/forschung/ressourcen/korpora/dire}.}
It is assembled on the basis of 2k object categories with 50
ImageNet\footnote{\url{http://imagenet.stanford.edu}} images each,
sampled from a larger dataset~\citep{Lazaridou:etal:2015b}. These are
natural images, which makes the task challenging. The object
categories given in the queries are those specified in ImageNet.

We build a set of linguistic attributes for each object by first
extracting the 500 most associated, and thus plausible, syntactic
neighbors for the category according to the DM
resource\ \citep{Baroni:Lenci:2010}. This excludes nonsensical
combinations such as \textit{repair:dog}. We further retain only
(relatively) abstract verbs taking the target item as direct
object.\footnote{We use the base form of verbs rather than the past
  participle for simplicity.} This is because (a)~concrete verbs are
likely to have strong visual correlates that could conflict with the
image (cf.~\emph{walk dog}); and (b)~referential expressions routinely
successfully mix concrete and abstract cues (e.g., \emph{the dog I
  own}). We remove all verbs with a score over 2.5 (on a 1--5 scale)
in the concreteness norms of \cite{brysbaert14:_concr_englis}.

We then construct each sequence as follows. First, we sample two
random categories, and three random entities (distinct images) for
each category (total: six entities). We then sample three attributes
compatible with both categories, giving us three attribute sets of
size two (a1+a2, a1+a3, a2+a3) to associate with the entities. We
create a completely balanced set of exposures by randomly pairing up
each of the three entities of each category with each of the three
attribute sets. Since this process gives us two exposures for each
entity (one with the first attribute, one with the second), it yields
a sequence of twelve exposures. The query is a random combination of a
category and two attributes, guaranteed to match exactly one entity.

% We also create extended versions of the dataset in which each exposure
% in an original sequence is repeated up to 3 or 5 times, respectively
% (for example, the first exposure from Fig.~\ref{fig:tracking-example}
% might occur 4 times, the second once, etc.). We consider this
% variation because one characteristic of entity tracking is that
% redundant information should not impact performance (once I have heard
% that a certain barkeeper was evaluated, hearing this information again
% should not impact my inferences about him).

%%% Local Variables:
%%% mode: latex
%%% TeX-master: "main-iwcs2017.tex"
%%% End:

\section{The DIRE Model}
\label{sec:model}

The core novelty of our model, DIRE (for DIstributed model of
REference), is a method to dynamically construct an \emph{entity
  library}, conceptually inspired in 1) the DRSs of DRT,\footnote{DRSs
  represent many types of information; as explained above, here we
  focus on entity-related information.} and 2)~\cite{joulin-mikolov15}
and \cite{graves+16}, who simulate discrete memory-building operations
in a differentiable continuous setup.\footnote{While we developed
  DIRE, \cite{henaff+16} proposed a similar architecture; we leave a
  comparison to future work.} The model is a feed-forward network
enhanced with a dynamic memory (the entity library), as well as
mechanisms to interact with it.

The entity library is updated after reading an input exposure by
either creating a new entity slot for the exposure, or adding the
exposure contents to an existing entity slot. This decision is based
on the similarity between the current input and the entities already
in the library. This generic mechanism
(Section~\ref{sec:gener-entity-libr}) can be applied in any setting
that accumulates information about entities over time. We explain how
we use it for our cross-modal tracking task in
Section~\ref{sec:cross-modal-tracking}.

\subsection{Building the DIRE Entity Library}
\label{sec:gener-entity-libr}

\begin{figure}[bt]
\centering
  \includegraphics[width=0.85\columnwidth]{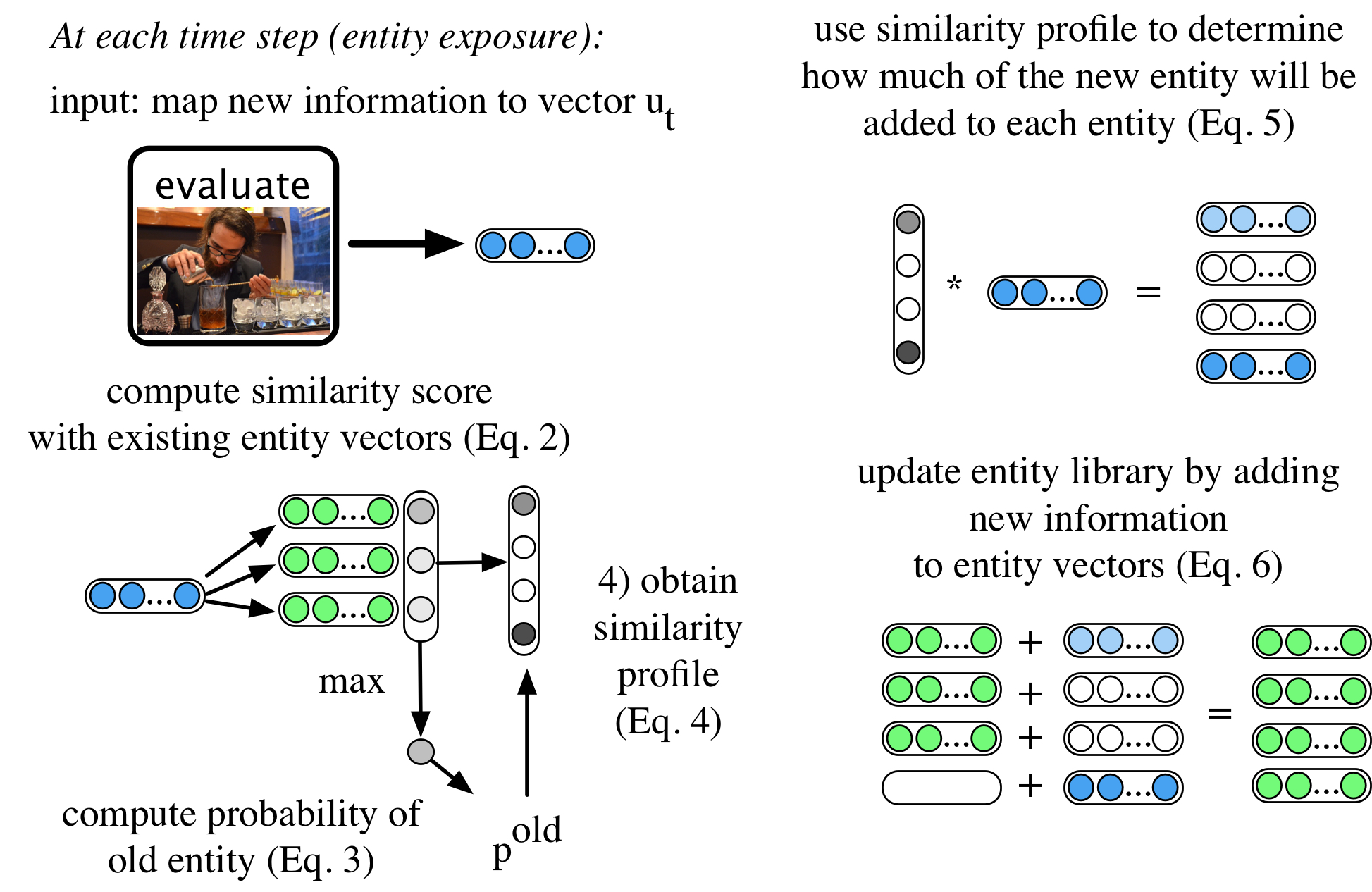}  
  \caption{Building the DIRE entity library.}
  \label{fig:dire-build}
\end{figure}

Figure~\ref{fig:dire-build} depicts the entity library building
mechanism. The input to the model is a set of subsequent exposures
$x_1, x_2, \dots{}, x_n$ which are represented by vectors
$\mathbf{u}_1, \mathbf{u}_2, \dots{}, \mathbf{u}_n$. At the $t$-th
exposure, the entity library is updated to state $\mathbf{E}_t$ as
follows. The first exposure vector $\mathbf{u}_1$ is added to the
entity library as is (Equation~\ref{eq:2}).  For $\mathbf{u}_{i>1}$,
we obtain a similarity profile $\mathbf{s}_i$ by taking its dot
product with the entity vectors in the library (Equation~\ref{eq:3};
note that $\mathbf{s}_i$ has $i-1$ dimensions):

\begin{eqnarray}
  \label{eq:2}
  \mathbf{E}_1=\mathbf{u}_1^\intercal\\
  \label{eq:3}
  \mathbf{s}_i = \mathbf{E}_{i-1} \mathbf{u}_i
\end{eqnarray}

The maximum similarity to an existing entity, $s_i^{max} =
\max(\mathbf{s}_i)$, cues whether $x_i$ is an instance of an entity
that has already been encountered before. We transform $s_i^{max}$
into $p^{old}_i$, the probability that exposure $x_i$ corresponds to
an ``old'' entity, as follows (with the scalar $w, b$ parameters
shared across all exposures for $i>1$):

\begin{equation}
  \label{eq:4}
  p^{old}_i = \sigma (w s_i^{max} + b)
\end{equation}

The entity library is updated by ``soft insertion''
\citep{joulin-mikolov15} of the current exposure vector $\mathbf{u}_i$
into the library. Concretely, we add the vector to each entity in the
library, weighted by the probability that the current exposure is an
instance of that entity. For the $i-1$ existing entities, this
probability is obtained by distributing the $p^{old}_i$ mass across
them, according to their probability of being the matching entity,
conditional on the exposure being old. The latter probability is
estimated by softmax-normalizing the similarity profile $\mathbf{s}_i$
from above. The probability that $x_i$ is new is obviously
$1-p^{old}_i$. This results in the following distribution, where $\|$
stands for concatenation:

\begin{equation}
  \label{eq:6}
  \mathbf{z}_i = p^{old}_i * \softmax (\mathbf{s}_i) \| ( 1 - p^{old}_i)
\end{equation}

Note that $z_i$ has one value more than the current number of stored
entities, expressing the probability that the current exposure
instantiates a new entity.
%  (including a 0 vector representing a new
% entity, see Eq.~\ref{eq:8}),

The entity library is then updated as:

\begin{eqnarray}
  \label{eq:7}%
  \mathbf{U}_i =  \mathbf{z}_i \mathbf{u}_i^{\intercal} \\
  \label{eq:8}%
  \mathbf{E}_i = (\mathbf{E}_{i-1} \|\mathbf{0}) + \mathbf{U}_i
\end{eqnarray}

Thus, we insert a 0 vector of the same dimensionality as the
$\mathbf{u}_i$ vectors at the end of the library, initializing a
blank slot to store a new entity. As a consequence, the library in its
end state will always contain as many entity vectors as
exposures. However, we expect those inserted for exposures of old
entities (that is, when $p^{old}_i \approx 1$) to be near zero, and
removable from the library along the lines
of~\cite{graves+16}.

\subsection{Cross-modal Entity Tracking with DIRE}
\label{sec:cross-modal-tracking}

\begin{figure}[bt]
\centering
  \includegraphics[width=0.75\columnwidth]{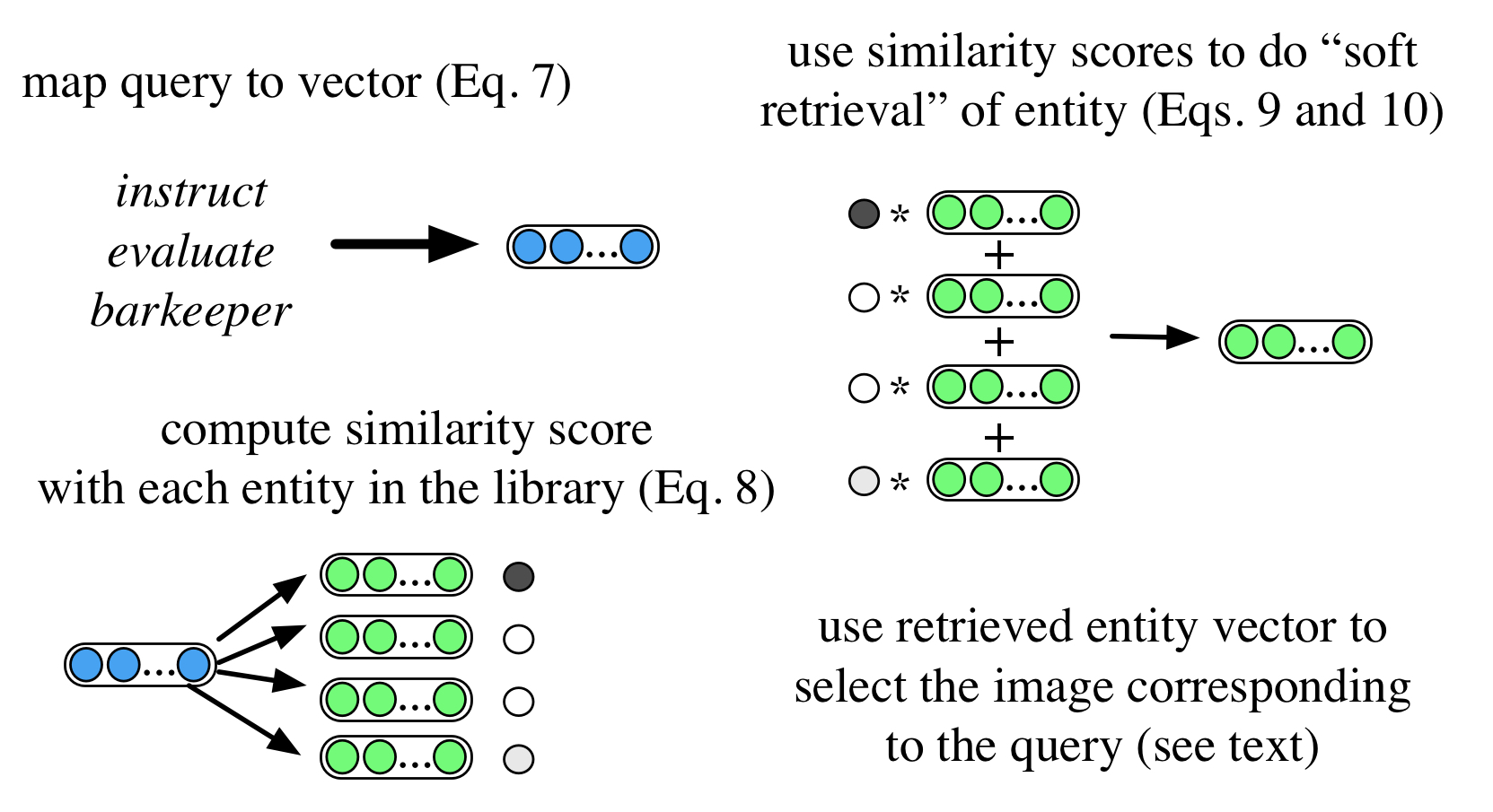}  
  \caption{Querying the DIRE entity library.}
  \label{fig:dire-query}
\end{figure}

We use DIRE for the cross-modal entity tracking task as follows (see
Figure~\ref{fig:dire-query}). Given pre-trained image and verbal
attribute representations, we first derive a multimodal representation
$\mathbf{u}_i$ for each exposure $x_i$ and update the entity library
as explained in Section~\ref{sec:gener-entity-libr}. The linguistic
query is mapped to the same multimodal space where entities live, and
the most relevant entity is retrieved. Finally, the images the model
has to choose from (candidate set) are also mapped to multimodal
space, and the correct answer is picked based on their similarity with
the retrieved entity.  We share the same $\mathbf{V}$ projection
across all images (in the exposures as well as in the candidate set at
query time), a single $\mathbf{A}$ projection for the verbal
attributes (in the exposures and in the query), and a matrix
$\mathbf{C}$ for the category name in the query. Details on each of
the steps follow.

\paragraph{Multimodal Mapping.}

Exposures are linearly mapped to a multimodal space combining visual
and linguistic information, building the $\mathbf{u}_i$ vector by
separately embedding each image vector $\mathbf{i}_i$ and attribute
vector $\mathbf{a}_i$ using a matrix $\mathbf{V}$ for
images\footnote{Size $v \times m$, where $v$ is the size of the image
  vector and $m$ the multimodal dimensionality.} and a matrix
$\mathbf{A}$ for linguistic attributes\footnote{Size $t \times m$,
  where $t$ is the size of the attribute vector. Both matrices, V and
  A, are learned.} and adding up the result
(Equation~\ref{eq:1a}). Storage takes place by feeding the
$\mathbf{u}_i$ vectors sequentially to the entity library
(cf.\ Section~\ref{sec:gener-entity-libr}).

\begin{equation}
  \label{eq:1a}
   \mathbf{u}_i = \mathbf{V} \mathbf{i}_i + \mathbf{A} \mathbf{a}_i
 \end{equation}

\paragraph{Query and Retrieval.} To select the best entity match for the
query, we compute a ``soft retrieval'' operation inspired by
\cite{sukhbaatar+15}. 
%We obtain a query-to-entity similarity profile
%by taking softmax-normalized dot products of the query vector with
%each vector in the entity library. We then sum the latter, weighting
%them by the corresponding scalar entries in the similarity
%profile. 
To query the entity library, we first map the query (a linguistic
referring expression, consisting of one noun and two attributes) to
multimodal space. We embed the attribute vectors $\mathbf{a}_1^q$, $\mathbf{a}_2^q$ with
the matrix $\mathbf{A}$ learned during storage and the noun vector $\mathbf{c}$ with
matrix $\mathbf{C}$, and we sum the result (Eq.~\ref{eq:1bis}). The query
vector $\mathbf{q}$ lives in the same space as the entity vectors, which
enables similarity computations.

\begin{equation}
  \label{eq:1bis}
  \mathbf{q} = \mathbf{C} \mathbf{c} + \mathbf{A} \mathbf{a}_1^q +
  \mathbf{A} \mathbf{a}_2^q
\end{equation}

We then retrieve the entity representation that matches the query, by
first computing the similarity of the mapped query $\mathbf{q}$ to
each entity vector $\mathbf{e}_i$ through a normalized dot product,
$\mathbf{g}$ (Eq.~\ref{eq:9}), and then using those similarities as
weights to perform a ``soft retrieval'' of the entity that best
matches the query, summing up the vectors in the entity library
multiplied by $\mathbf{g}$ (Eq.~\ref{eq:10}). Note that if only one
entity is significantly similar to the query (so that the
corresponding entry in the similarity profile tends to 1, while all
other entries tend to 0), this is equivalent to retrieving that
entity.

\begin{eqnarray}
  \label{eq:9}
  \mathbf{g} = \softmax (\mathbf{E}_{n} \mathbf{q}) \\
  \label{eq:10}
  \mathbf{r} = \mathbf{E}_n^\intercal \mathbf{g}
\end{eqnarray}

\paragraph{Picking the Right Image.} 
Finally, we use the retrieved entity representation, $\mathbf{r}$, to pick
among the $k$ images that represent the entities. We map the
candidate image vectors $\mathbf{d}_1 \dots \mathbf{d}_k$ to
multimodal space using the same visual matrix $\mathbf{V}$ as
above. We compare the query with each of the images using a dot
product, again obtaining a similarity profile, that we
softmax-normalize to obtain the final probability distribution that
will give us the candidate image, namely, the one corresponding to the
argmax of the probability distribution. Note that we need a
probability distribution because we use a cross-entropy cost function
when training the model.

The whole architecture is differentiable, allowing end-to-end training
by gradient descent; in particular, the cross-modal mapping is learned
as the model learns to refer.\footnote{Note that the input vectors for
  images are only visual, and those for nouns and attributes are only
  textual.} At the same time, it emulates discrete-like operations
like insertion and retrieval of entity representations, that, in
frameworks such as DRT, are performed entirely in symbolic terms, and
are manually coded in the DRT system Boxer~\citep{bos08}. This has the
advantage that the entity representations can be continuous, enabling
their matching with continuous representations of language as well as
cross-modal reasoning (for instance, using \textit{cup} for something
that a different speaker calls \textit{mug}, or mixing visual and
linguistically conveyed information). The model is rather
parsimonious, with parameters limited to three mapping matrices
($\mathbf{V}$, $\mathbf{A}$, $\mathbf{C}$) and the bias and weight
terms for $p^{old}$.

%%% Local Variables:
%%% mode: latex
%%% TeX-master: "main-iwcs2017.tex"
%%% End:

\section{Experiments}
\label{sec:experiments}

\paragraph{Experimental details.}
Images are represented by 4096-dimensional vectors produced by passing
images through the pre-trained VGG 19-layer CNN of
\cite{Simonyan:Zisserman:2015} (trained on the ILSVRC-2012 data), and
extracting the corresponding activations on the topmost fully
connected layer.\footnote{We use the MatConvNet toolkit,
  \url{http://www.vlfeat.org/matconvnet}.} Linguistic representations
are given by 400-dimensional \texttt{cbow} embeddings from
\cite{Baroni:etal:2014}, trained on about 2.8 billion tokens of raw
text. We map to a 1K-dimensional multimodal space. The parameters of
\textbf{DIRE} are estimated by stochastic gradient descent with 0.09
learning rate, 10 minibatch size, 0.5 dropout probability, and
maximally 150 epochs
\cite[here and below, hyperparameter values as in][]{baroni+17:show}.

As competitors, we train standard feed-forward (\textbf{FF}) and
recurrent (\textbf{RNN}) networks which have no external memory, using
two 300-dimensional hidden layers and sigmoid nonlinearities. We also
implement the related Memory Network model
\cite[\textbf{MemN};][]{sukhbaatar+15}. Like DIRE, MemN controls a memory structure,
but stores each input exposure separately in the memory. At the same
time, MemN can perform multiple ``hops'' at query time. Each hop
consists in soft-retrieving a vector from the memory, where the
probing vector is the sum of the input query vector and the vector
retrieved in the previous hop (null for the first hop). Conceptually,
DIRE attempts to merge different instances of the same entity at input
processing time, whereas MemN stores each piece of input separately
and aggregates relevant information at query time. MemN can thus use
the query to guide the search for relevant information. At the same
time, it does not optimize the way in which it stores information in
memory. Another difference with DIRE is that MemN uses \emph{two} sets
of mapping matrices: One to derive the vectors used at query time, the
other for the vectors used for retrieval. We employ the same
hyperparameters for MemN (also multimodal vector size) as for our model.

\paragraph{Results.}
Table~\ref{tab:results-tracking} shows that DIRE outperforms the
standard networks (FF and RNN) by a large margin, confirming the
importance of a discrete memory structure for reference tracking. If
we make the MemN architecture completely comparable to our model (with
one matrix and one hop, MemN-1m-1h), our model achieves higher results
(0.64 for DIRE-1m, 0.59 for MemN-1m-1h), which indicates that the
basic architecture of the model holds promise. However, MemN
outperforms DIRE when using two matrices, two hops (0.67
MemN-2m-1h/MemN-1m-2h vs.\ 0.65 DIRE-2m), or both (0.69
MemN-2m-2h). For MemN, this seems to be the upper bound, as increasing
to three hops greatly harms results (see last row).

Further analysis suggests that DIRE successfully addresses the two
challenges set out in the introduction: (i) It learns to
categorize: Only for 8\% of the datapoints does the model pick an
image of the wrong category, and these are cases where confounders
belong to visually similar or related categories to the target
(\textit{cottage-chalet}, \textit{youngster-enthusiast},
\textit{witch-potion}). It is worth noting that the model learns to
categorize directly from reference acts: At exposure time, the image
is not provided with a category label, so the model needs to induce
the category as part of solving the reference task. (ii) DIRE also
learns to individuate by combining visual and linguistically-conveyed
information: The similarity of the exposure to the query goes to
near-zero when the attribute is wrong, even when the category is the
same. Together, these two properties make it able to ground linguistic
expressions to entities represented in images. However, the entity
creation mechanism still needs to be fine-tuned, as currently DIRE
creates a new entity vector for nearly every exposure. More work is
needed for this crucial part of the model.

\begin{table}[t]
\centering
\small
% \begin{tabular}{ll}
%   Random baseline & 0.17\\
%   FF & 0.27\\
%   RNN & 0.28\\
%   % LSTM & 0.\\\hline
%   DIRE-1m & 0.64\\
%   DIRE-2m & 0.65\\
%   MemN-1m-1h & 0.59\\
%   MemN-1m-2h & 0.67\\
% %  MemN-2m-1h & 0.67\\ 
%   MemN-2m-2h & 0.69\\
%   % MemN-1m-3h & \textit{0.31} &   MemN-2m-3h & \textit{0.31}\\
% \end{tabular}
\begin{tabular}{llllllllllll}
  \toprule
  \multicolumn{2}{c}{Baseline} & 
  \multicolumn{2}{c}{Standard models} & 
  \multicolumn{4}{c}{DIRE} & 
  \multicolumn{4}{c}{MemN}  \\
  \cmidrule(r){1-2}
  \cmidrule(lr){3-4}
  \cmidrule(lr){5-8}
  \cmidrule(l){9-12} 
  Random & 0.17 & FF & 0.27 & 1m & 0.64 & 2m & 0.65 &
  1m-1h & 0.59 & 2m-1h & 0.67 \\
  & & RNN & 0.28 & & & & & 1m-2h & 0.67 & 2m-2h & \textbf{0.69} \\
  & & & & & & & & 1m-3h & 0.30 & 2m-3h & 0.30 \\
  \bottomrule
\end{tabular}
\caption{Tracking results (accuracy on test set).}
\label{tab:results-tracking}
\end{table}

%%% Local Variables:
%%% mode: latex
%%% TeX-master: "main-iwcs2017.tex"
%%% End:

\section{Discussion}
\label{sec:discussion}

% MemNN beat us but the hop parameter needs to be manually specified;
% our entity creation mechanism should remove the need to specify
% this, as we are abstracting away the information at input processing
% time such that it can be recovered in one go; however, this
% mechanism is still not working properly, so more work needs to be
% done

Providing a continuous model of reference that can emulate discrete
reasoning about entities is an ambitious research programme.  We have
reported on work in progress on such a model, DIRE, which, unlike
Memory Networks, and emulating formal approaches such as DRT within an
end-to-end neural architecture, is designed to make decisions as to
how to store the information \textit{at input processing time}, in a
way that aids further reasoning, namely, organizing it by
entity. Results suggest that merging complementary aspects of DIRE and
MemN could be fruitful. We have also presented a new task, cross-modal
entity tracking, that tests the categorization and individuation
capabilities of computational models, and a challenging dataset for
the task.

Our project is related to several areas of active research. Reference
is a classic topic in philosophy of language and
linguistics~\citep{frege1892,abbott10,kamp-reyle93,kamp15}; emulating
discrete aspects of language and reasoning through continuous means is
a long-standing goal in artificial
intelligence~\citep{Smolensky:1990,joulin-mikolov15}, and recent work
focuses on
reference~\citep{baroni+17:show,herbelot:2015:IWCS2015,herbelot-vecchi:2015:EMNLP};
grounding language in
perception~\citep{chen-mooney11,bruni+12:ACL,silberer+13}, as well as
reference and co-reference
\citep{krahmer-van-deemter12,poesio+17:anaphora_book} are important
subjects in Computational Linguistics. Our programme puts these
different strands together.

% In future work we will concentrate on 1)~deepening the connections to
% the aforementioned research lines, 2)~improving the model, in
% particular merging complementary aspects of DIRE and MemN, and
% 3)~addressing more realistic data.

%%% Local Variables:
%%% mode: latex
%%% TeX-master: "main-iwcs2017.tex"
%%% End:

\paragraph*{Acknowledgments:}
We thank Angeliki Lazaridou for help producing the visual vectors used
in the paper.
%GBT: They require this exact wording:
This project has received funding from the European Research Council
(ERC) under the European Union's Horizon 2020 research and innovation
programme (grant agreement No 715154; AMORE); EU Horizon 2020
programme under the Marie Sk\l{}odowska-Curie grant agreement No 655577
(LOVe); ERC 2011 Starting Independent Research Grant n.~283554
(COMPOSES); DFG (SFB 732, Project D10). We also gratefully acknowledge
the support of NVIDIA Corporation with the donation of GPUs to
U. Trento and U. Pompeu Fabra.
%GBT: and the logo:
%GBT: and also the disclaimer:
This paper reflects the authors' view only, and the EU
is not responsible for any use that may be made of the information it contains.\begin{flushright}
\includegraphics[width=0.8cm]{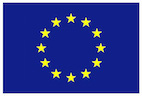}
\includegraphics[width=0.8cm]{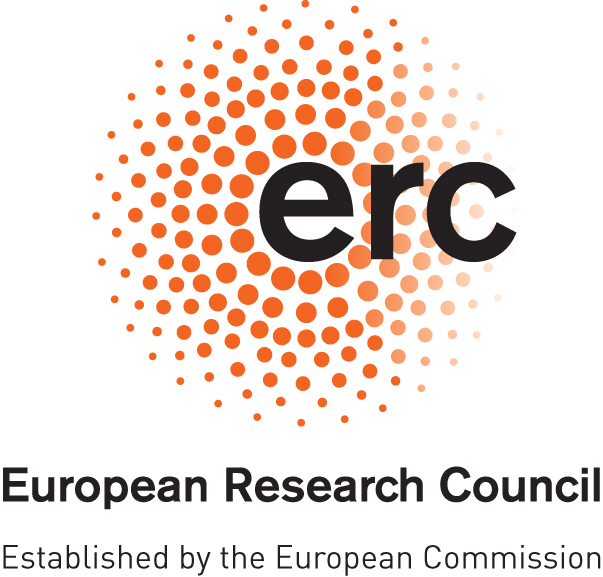} 
\end{flushright}
%GBT: Sigh.

\bibliographystyle{chicago}
\bibliography{marco,gemma}

%\appendix

\end{document}